\documentclass[letterpaper, 10 pt, conference]{ieeeconf}  

\IEEEoverridecommandlockouts                              

\usepackage{multirow}
\usepackage{amsmath}
\usepackage{booktabs}
\usepackage{graphicx}
\usepackage{hyperref}

\hypersetup{hypertex=true,
            colorlinks=true,
            linkcolor=black,
            anchorcolor=black,
            citecolor=black}

\usepackage{bm}
\usepackage{algorithmicx,algorithm}
\usepackage[noend]{algpseudocode}

\title{\LARGE \bf
Efficient Trajectory Planning and Control for USV with Vessel Dynamics and Differential Flatness
}

\author{Tao Huang$^{1,2}$, Zhenfeng Xue$^{1,2,*}$, Zhe Chen$^{1,2}$, Yong Liu$^{1,2,*}$
\thanks{$^{1}$Tao Huang, Zhenfeng Xue, Zhe Chen and Yong Liu are with Institute of Cyber-Systems and Control, Zhejiang University, Hangzhou, China}%
\thanks{$^{2}$Tao Huang, Zhenfeng Xue, Zhe Chen and Yong Liu are Intelligent Perception and Control Center, Huzhou Institute of Zhejiang University, Huzhou, China}%
\thanks{$^*$Corresponding: Zhenfeng Xue (zfxue0903@zju.edu.cn), Yong Liu (yongliu@iipc.zju.edu.cn)}
}

\begin{document}

\maketitle
\thispagestyle{empty}
\pagestyle{empty}

\begin{abstract}

Unmanned surface vessels (USVs) are widely used in ocean exploration and environmental protection fields. To ensure that USV can successfully perform its mission, trajectory planning and motion tracking are the two most critical technologies.
In this paper, we propose a novel trajectory generation and tracking method for USV based on optimization theory.
Specifically, the USV dynamic model is described with differential flatness,  so that the trajectory can be generated by dynamic RRT$^*$ in a linear invariant system expression form under the objective of optimal boundary value. 
To reduce the sample number and improve efficiency, we adjust the trajectory through local optimization.
The dynamic constraints are considered in the optimization process so that the generated trajectory conforms to the kinematic characteristics of the under-actuated hull, and makes it easier to be tracked.
Finally, motion tracking is added with model predictive control under a sequential quadratic programming problem.
Experimental results show the planned trajectory is more in line with the kinematic characteristics of USV, and the tracking accuracy remains a higher level.

\end{abstract}

\section{Introduction}

Unmanned surface vehicles (USVs) are designed to relieve human labor in various water surface missions, such as surface cleaning, cyanobacteria treatment, carga transportation and military reconnaissance.
It mainly perceives and maps the surrounding environment through shipborne radar or camera, and then performs path planning and motion control automatically by the core controller~\cite{niu2019voronoi,shah2019long,chen2021novel}.
In the travelling process, the USV needs to accurately avoid various static and dynamic obstacles in the environment.
To ensure that USV can successfully complete its mission autonomously, obstacle avoidance trajectory planning~\cite{wen2020collision} con control~\cite{han2021efficient} are the two most critical technologies.

However, current trajectory planning algorithms are usually migrated directly from other similar unmanned platforms, such as unmanned aerial vehicle (UAV) or automated guided vehicle (AGV), which ignores the inherent kinematic characteristics of USV and thus degrades the motion tracking control performance.
Specifically, the USV is a classical under-actuated system, which means the number of inputs is less than the actual degree of freedom of the system.
We usually regard the USV as a system with a three degree of freedom, \textit{i.e.}, surge, sway and yaw, but the drive dimension only includes propeller thrust along surge axis and its torque along yaw axis.
This makes the generated path hard to be tracked for USV due to the lack of relevance between path planning and tracking. 

\begin{figure}[t]
	\centering
	\includegraphics[scale=0.35]{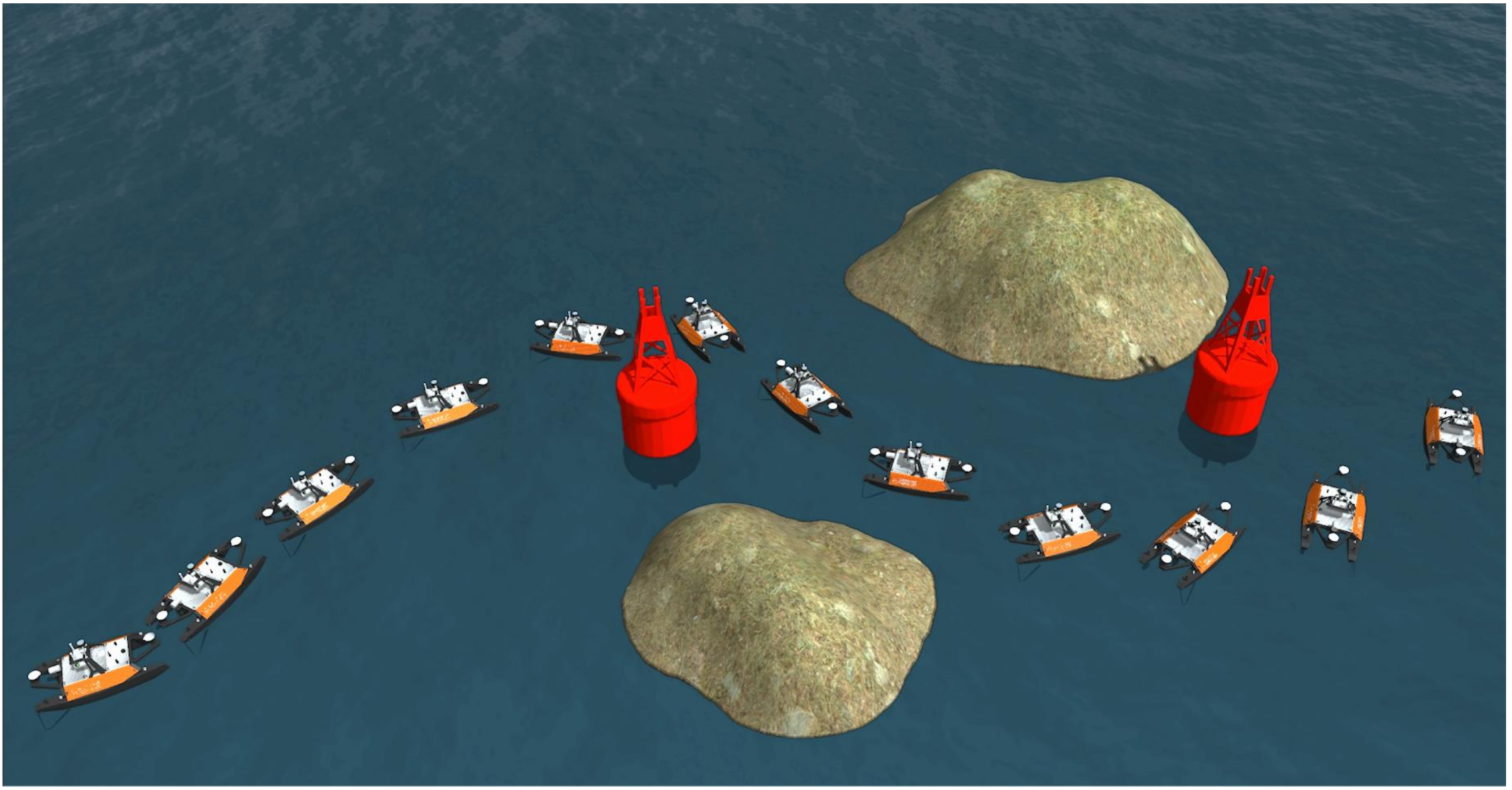}
	\caption{A composite image of USV sailing in obstacle environment with planned trajectory that fits the dynamic characteristics.}
	\label{fig:plan_result}
\end{figure}

In this paper, we propose a novel trajectory planning and control method specifically designed for USV.
Under the circumstance that the global map is assumed to be known, the dynamic RRT$^*$ algorithm is firstly applied to generate the initial obstacle avoidance trajectory.
In the process of constructing the sampling interval, we propose to combine the Dubins curve~\cite{dubins1957curves} and A$^*$ algorithm, because the smooth and evasive characteristics of Dubins curve fits the hull dynamics well.
After that, the path planning problem is expressed as a linear time invariant (LTI) form, and an trajectory optimization is established while concerning the spatial quality of the path, such as smoothness, obstacle avoidance and vessel dynamics.
Different from the existing methods~\cite{wen2020collision,du2019trajectory} that directly use the dynamic model of USV system as the equality constraint in the trajectory generation process, so that the generated trajectory satisfies the motion characteristics of the USV, which causes huge computational burden and unstable generation results.
Here, we use differential flatness to describe the USV dynamic system to solve the optimization problem.
Our proposed method greatly improves the optimization efficiency, and the usage of differential flatness can connect path planning with motion control well, and make the subsequent motion tracking control more accurate.

The effect of the proposed path planning algorithm is illustrated in Fig.~\ref{fig:plan_result}, from which the USV is able to travel across the map of multiple obstacles efficiently.
The generated path is smooth enough such that it can be well consistent with the motion characteristics of the USV.
In the subsequent motion control process, we successfully combine the differential flatness of system with the nonlinear model predictive control (NMPC).
A quadratic optimization problem about the error between given trajectory state and the forward prediction state is constructed and cyclically solved.
The tracking accuracy remains at a high level during the rapid operation of the USV no matter whether there is directional water interference.

\section{Related works}

\subsection{Path planning for USV}

It is generally based on some optimization objectives for USV trajectory planning, such as the shortest distance, the least time or the best energy consumption, to stably generate feasible and safe routes.
Similar to other unmanned platforms, the motion planning algorithm for USV includes two key contents, \textit{i.e.}, path generation and obstacle avoidance.
Path generation makes the USV to quickly acquire the feasible route to the mission target based on the map information.
Wang \textit{et al.}~\cite{wang2019multilayer} proposed a multi-layer path planner for route generation and collision avoidance based on fast marching method (FMM) and B-spline. It not only ensures the planning speed, but also smooths the path due to obstacle avoidance and improves the safety of USV navigation.
In addition to the traditional graph search methods~\cite{niu2019voronoi,yu2021usv,shah2019long}, there are also relevant applications based on reinforcement learning~\cite{guo2020autonomous} in the same problems.
Wen \textit{et al.}~\cite{wen2020collision} proposed a hierarchical trajectory planning that constructs an optimization problem based on RRT$^*$ algorithm and dynamic model of USV.
Shah \textit{et al.}~\cite{shah2014trajectory} as well as Du \textit{et al.}~\cite{du2019trajectory} modelled the problem in the state space and proposed to sample a reasonable trajectory based on motion primitives.
In order to effectively describe obstacle information, various improved artificial potential field (APF) methods~\cite{xie2014obstacle,chen2021path} are used in USV obstacle avoidance strategy, and obstacle environment description based on safety corridor can be used in complex water environment.
Compared with above methods, this paper proposes an USV trajectory optimization algorithm that combines the Dubins curve~\cite{dubins1957curves} and the dynamic RRT$^*$ to improve the sampling efficiency.
While stably generating the navigation trajectory that meets the motion characteristics of USV, it ensures the real-time performance and quick response to unknown obstacles.

\subsection{Motion control for USV}

After the feasible navigation trajectory is obtained, robust trajectory tracking is another key component to realize the autonomous navigation of USV.
Due to the uncertain external disturbance, the travel trajectory of USV without reliable control will be greatly different from the expected navigation trajectory.
In the field of USV motion control, different control methods have been proposed, such as back stepping~\cite{wen2018adaptive}, sliding mode control~\cite{zhang2020nonsingular}, adaptive control~\cite{dong2020neural} and model predictive control~\cite{abdelaal2018nonlinear,wang2020roboat}.
These control methods deal well with the problem of robust tracking of the USV path in the presence of external disturbances.
However, these works generally only focus on the tracking of the USV trajectory, or simply consider the planning task of USV path, but they ignore whether the generated path can be directly used for tracking control.
In the relevant USV path tracking literature~\cite{han2021efficient}, the kinematic model of USV is added to path generation to make the navigation more suitable for USV tracking control, but the computational burden and algorithm stability become tough problems.
Referring to~\cite{sreenath2013geometric,morales2015linear,zhou2019robust}, this paper proposes the differential flatness property of the USV motion system to simplify the kinematic model as an algebraic expression of the position and its higher order derivatives.
It effectively connects the navigation motion planning with the USV tracking control problem, so that the USV can more stably follow the trajectory and improve the stability and successful rate.

\section{Description of USV system}

An USV is an under-actuated system with six degrees of freedom. When its dynamic characteristics are directly embedded into path planning and motion control, the results could be unstable. One feasible solution is to use flatness to describe the system.

\subsection{USV discrete model}

\begin{figure}[t]
	\centering
	\includegraphics[scale=0.6]{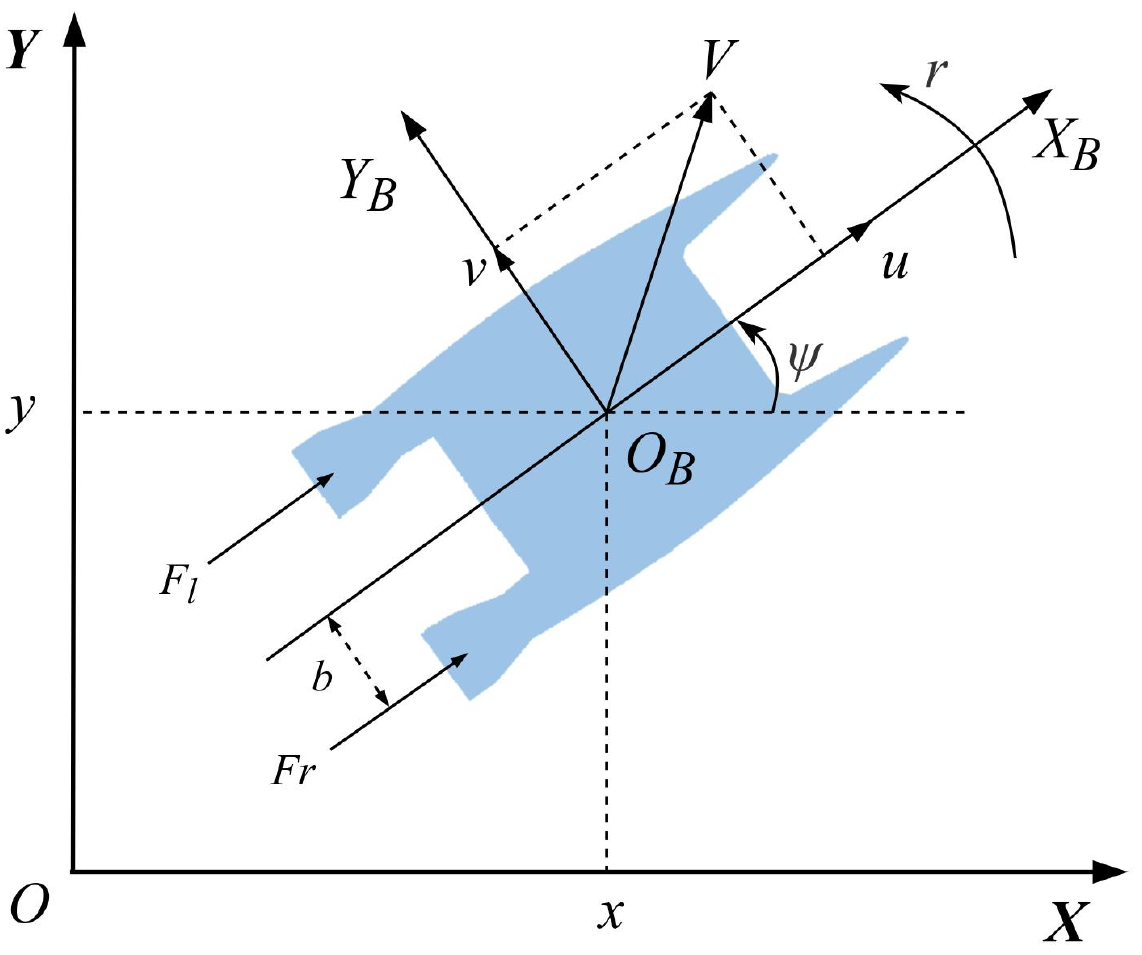}
	\caption{USV earth-fixed and body-fixed coordinate frames.}
	\label{fig:usv_dynamics}
\end{figure}

As shown in Fig.~\ref{fig:usv_dynamics}, an USV with kinematics and dynamics in earth-fixed and body-fixed frames can be modelled by
\begin{equation}\label{equ:usv_model}
	\left\{
	\begin{aligned}
		\dot{\bm\eta} &= \bm{J}(\bm\eta) \bm\nu \\
		\bm{M}\dot{\bm\nu} &= \bm\tau -\bm{C}(\bm\nu)\bm\nu-\bm{D}\bm\nu,
	\end{aligned}
	\right.
\end{equation}
where $\bm\eta=[x, y, \varphi]^T$ and $\bm\nu=[u,v,r]^T$ are position, heading and velocity vectors in the earth-fixed and body-fixed frames respectively. $\bm{J}(\bm\eta)$ is the rotation matrix, and $\bm{M}$ is positive-definite inertia matrix. $\bm{C}(\bm\nu)$ is the Coriolis and centripetal force matrix, and $\bm{D}$ is hydrodynamic damping matrix. $\bm\tau=[\tau_u,\tau_v,\tau_r]^T$ is the thrust matrix.

For simplicity, considering a catamaran, we suppose the hull is symmetrical about the $u$-axis and $v$-axis. The constraint of $v$-axis symmetry is over strict, but practice has proved it is almost feasible for a catamaran. Thus, the relevant matrix can be expressed as
\begin{equation}
	\bm{M} = diag\{ m_1, m_1, m_2 \},
\end{equation}
\begin{equation}
	\bm{C}(\bm\nu) = \begin{bmatrix}
		0 & 0 & -m_2v	\\
		0 & 0 & m_1u \\
		m_2v & -m_1u & 0
	\end{bmatrix},
\end{equation}
\begin{equation}
	\bm{D} = diag\{ d_1, d_2, d_3 \},
\end{equation}
and the thrust matrix can be expressed as
\begin{equation}
	\bm\tau = \begin{bmatrix}
		F_{r}+F_{l} \\
		0\\
		b \cdot (F_{r}-F_{l})
	\end{bmatrix}.
\end{equation}

Combined with the above equations, the USV dynamic model is rewritten as the form of state equation as follow
\begin{equation}
	\dot{\bm{x}}(t)=f(\bm{x}(t), \bm{u}(t)),
\end{equation}
where the state and input of system is $\bm{x}=[x,y,\varphi,u,v,r]^T$ and $\bm{u}=[\tau_u, \tau_r]^T$ respectively.

In order to reflect the motion characteristics of USV in discrete time domain, we use the explicit 4th order Runge Kutta method to calculate the form
\begin{equation}
	\bm{x}_{i+1}(t)=f_{RK4}(\bm{x}_{i}(t), \bm{u}_{i}(t), \varDelta{t}),
\end{equation}
where $\varDelta{t}$ is the discrete step. There are unknown hydrodynamic parameters $\bm{h}=[m_1, m_2, d_1, d_2, d_3]$ in the above USV system model, which will be used in subsequent planning and control methods. Based on Eq.~\ref{equ:usv_model}, we can use collected hull real motion state data $v_r$ and $u_r$ to estimate $\bm{h}$ with the following optimization problem
\begin{equation}
	\begin{gathered}
		\min _{\bm{h}} \sum_{t=0}^{T_{r}} \bm\varepsilon(t)^{T} \bm\Omega \bm\varepsilon(t) \\
		s.t. \quad \bm{h}_{l} \leq \bm{h} \leq \bm{h}_{u} \\
		\bm{x}_{i+1}(t)=f_{RK4}(\bm{x}_{i}(t), \bm{u}_{i}(t), \varDelta{t}, \bm{h}), t \in [0,T_r],
	\end{gathered}
\end{equation}
where $\bm\varepsilon(t)$ is the deviation between hull real state data $v_e$ and theoretical data $v$, and $\bm\Omega$ is weight matrix. The solution can be solved by Gauss Newton Hessian approximation method with the help of Casadi~\cite{andersson2019casadi} optimization library.

\subsection{Differential flatness}~\label{sec:flatness}

Due to the USV system is under-actuated and has nonholonomic constraints, real-time planning and accurate trajectory control exist major challenges. Wen \textit{et al.}~\cite{wen2020collision} try to insert the dynamic model of USV system as the equality constraint into trajectory generation, so that the generated trajectory meets the motion characteristics of USV. However, this leads to heavy computational burden and unstable results. Similar to~\cite{fliess1995flatness}, it is feasible to simplify the problem that USV model constraints need to be considered in trajectory generation by using the flat property of the system.

According to~\cite{fossen1999guidance}, the USV system is controllable and also has differentiable flatness. Then the system has a flat output $\bm{z}(t) \in \mathcal{R}^m$, so that the state and input can be expressed in the algebraic form of derivatives using only $\bm{z}(t)$ and its finite order~\cite{morales2015linear}, that is
\begin{equation}\label{equ:system_flatness}
	\left\{
	\begin{aligned}
		\bm{x}(t)&=\phi_{0}(\bm{z}(t),\dot{\bm{z}(t)},\ddot{\bm{z}(t)},\bm{z}^{(3)}(t)),\\
		\bm{u}(t)&=\phi_{1}(\bm{z}(t),\dot{\bm{z}(t)},\ddot{\bm{z}(t)},\bm{z}^{(3)}(t),\bm{z}^{(4)}(t)),
	\end{aligned}
	\right.
\end{equation}
where the flat output can be selected in USV system is $\bm{z}(t) = [x(t), y(t)]$. Then the state space of USV trajectory generation can be transformed into the consideration of state $[x,y]$ and its derivatives. The highest derivative order of $\bm{z}(t)$ in Eq.~\ref{equ:system_flatness} is 4. That is to say, the trajectory equation needs to reach at least 5 orders to ensure smoothness.

\section{Trajectory generation}

In complex water environment, USV trajectory generation faces the problems of real-time and feasibility of trajectory execution. Based on the optimized dynamic RRT$^*$ framework, we are able to generate trajectory that meets the execution of USV. Then the sampling state optimization method is added to the sampling tree expansion process to improve its speed.

\subsection{Path searching}

\begin{figure}[t]
	\centering
	\includegraphics[scale=0.42]{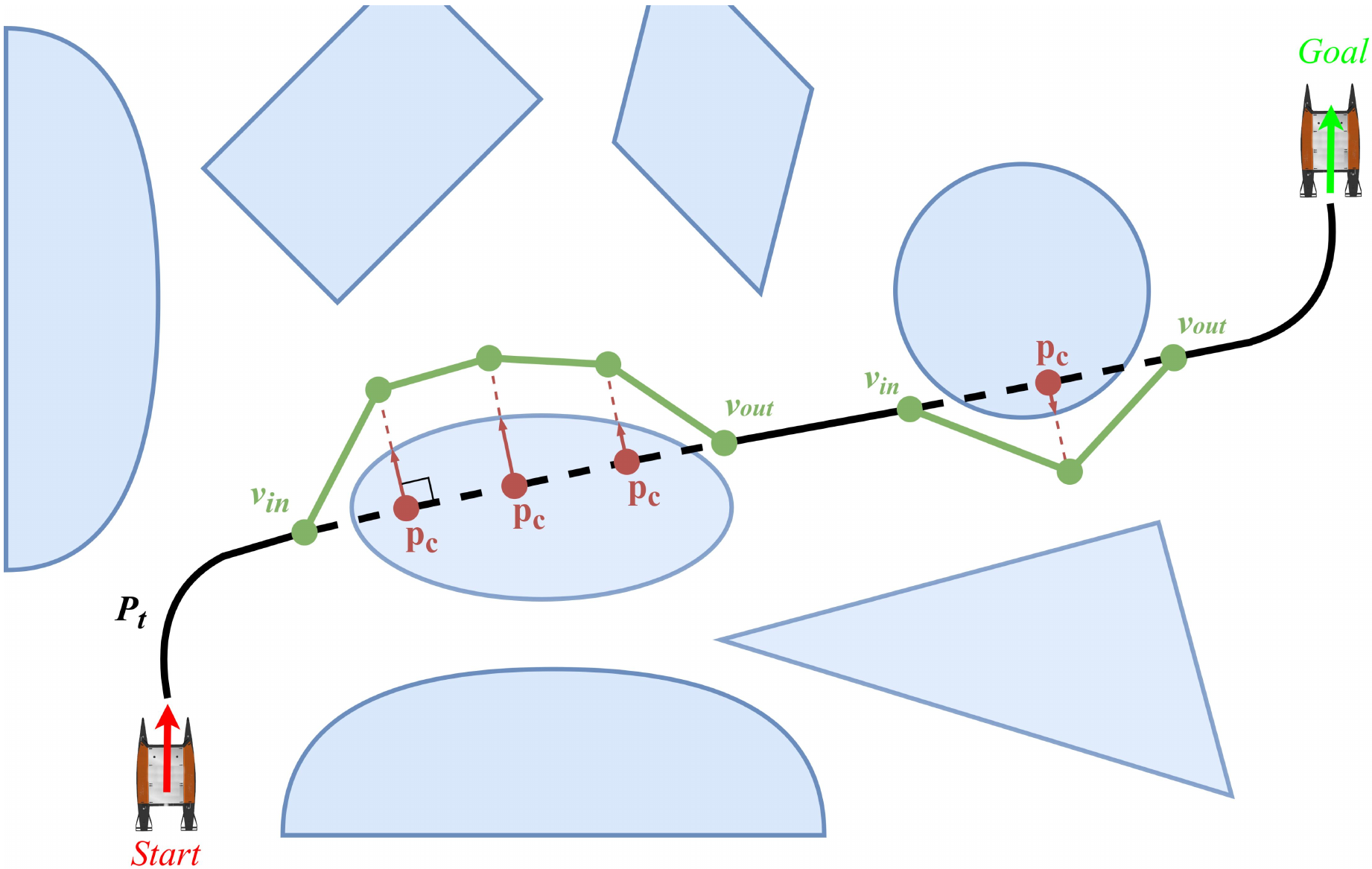}
	\caption{An illustration of the path search process. The black path obtained by Dubins curve has several collision points $p_c$, which will be moved outside the collision with a safe distance, resulting the green path.}
	\label{fig:path_search_illu}
\end{figure}

\begin{algorithm}[t]
	\caption{Kinodynamic Trajectory Plan} 
	\hspace*{0.02in} {\bf Input:} 
	Environment $e$, SamplingTree $\tau$, State $\bm{x}$\\
	\hspace*{0.02in} {\bf Output:}  
	Trajectory $T$
	\begin{algorithmic}[1]
		\State Initialize: $\tau_s\leftarrow\bm{x}_{start}, \tau_g\leftarrow\bm{x}_{goal}$
		\State $\mathcal{P}_t\leftarrow\bm{TopoPathFind}(e)$
		\For{i = 1 to n} 
			\State $\bm{x}_{rand}\leftarrow\bm{SampleState}(e,\bm{p}_t)$
			\State $\bm{x}_{front},\bm{x}_{back}\leftarrow\bm{NeighborFind}(\tau_s,\tau_g,\bm{x}_{rand})$
			\State $\bm{x}_s\leftarrow\bm{TreeGrow}(\bm{x}_{back},\bm{x}_{rand})$
			\If{$\bm{x}_s$ not empty}
				\State $\tau_s\leftarrow\tau_s\cup {\bm{x}_{rand},\bm{x}_s}$
			\EndIf
			\State $\bm{x}_g\leftarrow\bm{TreeGrow}(\bm{x}_{front},\bm{x}_{rand})$
			\If{$\bm{x}_g$ not empty}
				\State $\tau_g\leftarrow\tau_g\cup {\bm{x}_{rand},\bm{x}_g}$
			\EndIf
			\State $\bm{Rewire}(\tau_s,\tau_g,\bm{x}_{rand})$
			\If{$\bm{x}_{rand}\in\tau_s \wedge\bm{x}_{rand}\in\tau_g$}
				\State get one solution
			\EndIf
		\EndFor
		\State $T\leftarrow\bm{Recall}(\tau_s,\tau_g)$
		\State $\bm{TrajectoryOptimize}(T)$
		\State \Return $T$
		
	\end{algorithmic}
\end{algorithm}

The main process of dynamic RRT$^*$ framework is shown in Algorithm 1. By expanding two sampling trees $\tau_s$ and $\tau_g$, feasible trajectory from $\bm{x}_{init}$ to $\bm{x}_{goal}$ is searched rapidly.
In order to express the characteristics of USV nonholonomic constraints in the trajectory and improve the efficiency and stability of sampling, we combine the Dubins~\cite{dubins1957curves} and A$^*$ algorithm to generate the key topological path point $\mathcal{P}_t$, which is illustrated in Fig.~\ref{fig:path_search_illu}.
The Dubins curve is used here due to its smoothness, which is very suitable for the characteristics of under-actuated system.
Based on this, the sampling interval is constructed to effectively reduce the sampling of irrelevant areas.
The $\bf{Steer}()$ function in the $\bf{TreeGrow}()$ and $\bf{Rewire}()$ steps constructs the transfer trajectory from $\bm{x}_{rand}$ to sampling tree $\tau_s$ and $\tau_g$ that meets the USV dynamics.
In the next part, we will introduce to use closed form minimal value cost function to ensure stability and efficiency. 
If the sampling status $\bm{x}_{rand}$ exists in the sampling tree $\tau_s$ and $\tau_g$ at the same time, then a feasible trajectory is searched.

In the traditional RRT$^*$ framework, when the sampling state cannot be connected to the sampling tree due to obstacles or dynamic constraints, the discarding method is adopted. This increases the time-consuming of the algorithm.
To reduce the occurrence of this situation, we use optimization method to adjust the inappropriate transfer trajectory locally, so that the trajectory meets the feasibility requirements.

\subsection{Trajectory planning with flatness}

In section~\ref{sec:flatness}, the differentiable flatness of USV has been described. Based on this system characteristics, we only need to pay attention to the flat output $\bm{z}(t)$ and its higher derivatives when considering the planning state space of trajectory.
Here, we use linear time invariant (LTI) system to express the motion characteristics of the system, and the trajectory can be denoted as a parametric polynomial about time.
The state equation of USV system is expressed as
\begin{equation}
	\begin{gathered}
		\dot{\bm{x}}(t)=\bm{A} \bm{x}(t)+\bm{B u}(t),\\
		\bm{A}=\left[\begin{array}{ll}
			\bm{0} & \bm{I} \\
			\bm{0} & \bm{0}
		\end{array}\right], \quad \bm{B}=\left[\begin{array}{l}
			\bm{0} \\
			\bm{I}
		\end{array}\right],\\
		\bm{x}(t)=\left[ p(t), \dot{p}(t), \ddot{p}(t)\right]^{\top}, \quad \bm{u}(t)=\dddot{p}(t).
	\end{gathered}
\end{equation}
Therefore, the goal of trajectory generation is to calculate a cubic differentiable trajectory polynomial.
The aim is to guide the USV from initial state $\bm{x}_{init}$ to final state $\bm{x}_{end}$ along the time, \textit{i.e.}, the optimal boundary value problem (OBVP).
Similar to~\cite{zhou2019robust}, the following cost function needs to be optimized.
\begin{equation}~\label{eq:cost_function}
 		\mathcal{J}_{seg}=\int_{0}^{T}\left(\sigma+\frac{1}{2}\|\bm{u}(t)\|^{2}\right)dt,
\end{equation}
where $\sigma$ denotes the weight of time.
Based on the cost Eq.~\ref{eq:cost_function}, we can obtain the expression of the optimal trajectory from $\bm{x}_{init}$ to $\bm{x}_{end}$ with a closed form solution by using Pontryagin minimization method.
\begin{equation}
	\begin{gathered}
		\bm{x}^{*}(T)=\left[\begin{array}{c}
			\frac{\alpha}{120} T^{5}+\frac{\beta}{24} T^{4}+\frac{\gamma}{6} T^{3}+\frac{a_{0}}{2} T^{2}+v_{0} T+p_{0} \\
			\frac{\alpha}{24} T^{4}+\frac{\beta}{6} T^{3}+\frac{\gamma}{2} T^{2}+a_{0} T+v_{0} \\
			\frac{\alpha}{6} T^{3}+\frac{\beta}{2} T^{2}+\gamma T+a_{0}
		\end{array}\right]\\
		\bm{u}^{*}(T)=\frac{\alpha}{2}T^2+\beta T+\gamma,
	\end{gathered}
\end{equation}
where the parameters $\alpha$, $\beta$ and $\gamma$ can be expressed by $\bm{x}_{init}$ and $\bm{x}_{end}$, referring to~\cite{mueller2015computationally}.
According to the form of $\bm{u}^*(T)$, Eq.~\ref{eq:cost_function} can be simplified as
\begin{equation}
	\begin{gathered}
	\mathcal{J}_{seg}^{*} = \sigma T+\sum_{i\in{x,y}}(\frac{\alpha_{i}^2}{20}T^5+\frac{\alpha_{i}\beta_{i}}{4}T^4+\frac{\alpha_{i}\gamma_{i}+\beta_{i}^2}{3}T^3\\
	+\beta_{i}\gamma_{i} T^2+\gamma_{i}^2T).
\end{gathered}
\end{equation}
It can be found that $J^*_{seg}$ is a higher-order polynomial about time $T$.
Then the optimal trajectory can be constructed by calculating the minimum cost time by solving $\mathcal{J}^*_{seg}=0$.

The $\bf{TreeGrow}()$ process is illustrated in Algorithm 2, where $\bf{Steer}()$ quickly obtains the optimal trajectory between two states by solving Eq.~\ref{eq:cost_function} in a closed form, so that the algorithm can obtain the feasible trajectory.
The obtained trajectory can be added into the sampling tree only when it passes the feasibility test of $\bf{FeasibilityCheck}()$.

\begin{algorithm}[t]
	\caption{$TreeGrow$} 
	\hspace*{0.02in} {\bf Input:} 
	Environment $e$, SamplingTree $\tau$, \\
	NeighborState $\bm{x}_{n}$ SamplingState $\bm{x}_{rand}$\\
	\hspace*{0.02in} {\bf Output:}  
	ParentState $\bm{x}_{parent}$
	\begin{algorithmic}[1]
		\State Initialize: $\bm{x}_{parent}\leftarrow null,cost_{min}\leftarrow \infty$
		\For{$\bm{x}_cur$ in $\bm{x}_{n}$} 
		\State $T_{seg}\leftarrow\bm{Steer}(\bm{x}_cur,\bm{x}_{rand})$
		\If{$\bm{FeasibilityCheck}(T_{seg},e)$}
		\If{$\bm{TrajCost}(T_{seg}) < cost_{min}$}
		\State $\bm{x}_{parent}\leftarrow\bm{x}_cur,$
		\State $cost_{min}\leftarrow\bm{TrajCost}(T_{seg})$
		\EndIf
		\Else
		\State $T_{seg}\leftarrow\bm{TrajectoryOptimize}(T_{seg},e)$
		\If{$\bm{FeasibilityCheck}(T_{seg},e)$}
		\State $\bm{x}_{parent}\leftarrow\bm{x}_cur,$
		\State $cost_{min}\leftarrow\bm{TrajCost}(T_{seg})$
		\EndIf
		\EndIf
		\EndFor
		\State \Return $\bm{x}_{parent}$
	\end{algorithmic}
\end{algorithm}

\section{Trajectory optimization and tracking}

In the framework of trajectory generation, the feasibility of trajectory can not be guaranteed.
To reduce the number of samples and improve efficiency, we adjust the trajectory to meet the relevant constraints through local optimization according to the infeasible trajectory.
Referring to~\cite{mellinger2011minimum}, the closed form of solution can be calculated by setting the objective function to a quadratic form.

\subsection{Optimization modelling}

\begin{figure}[t]
	\centering
	\includegraphics[scale=0.4]{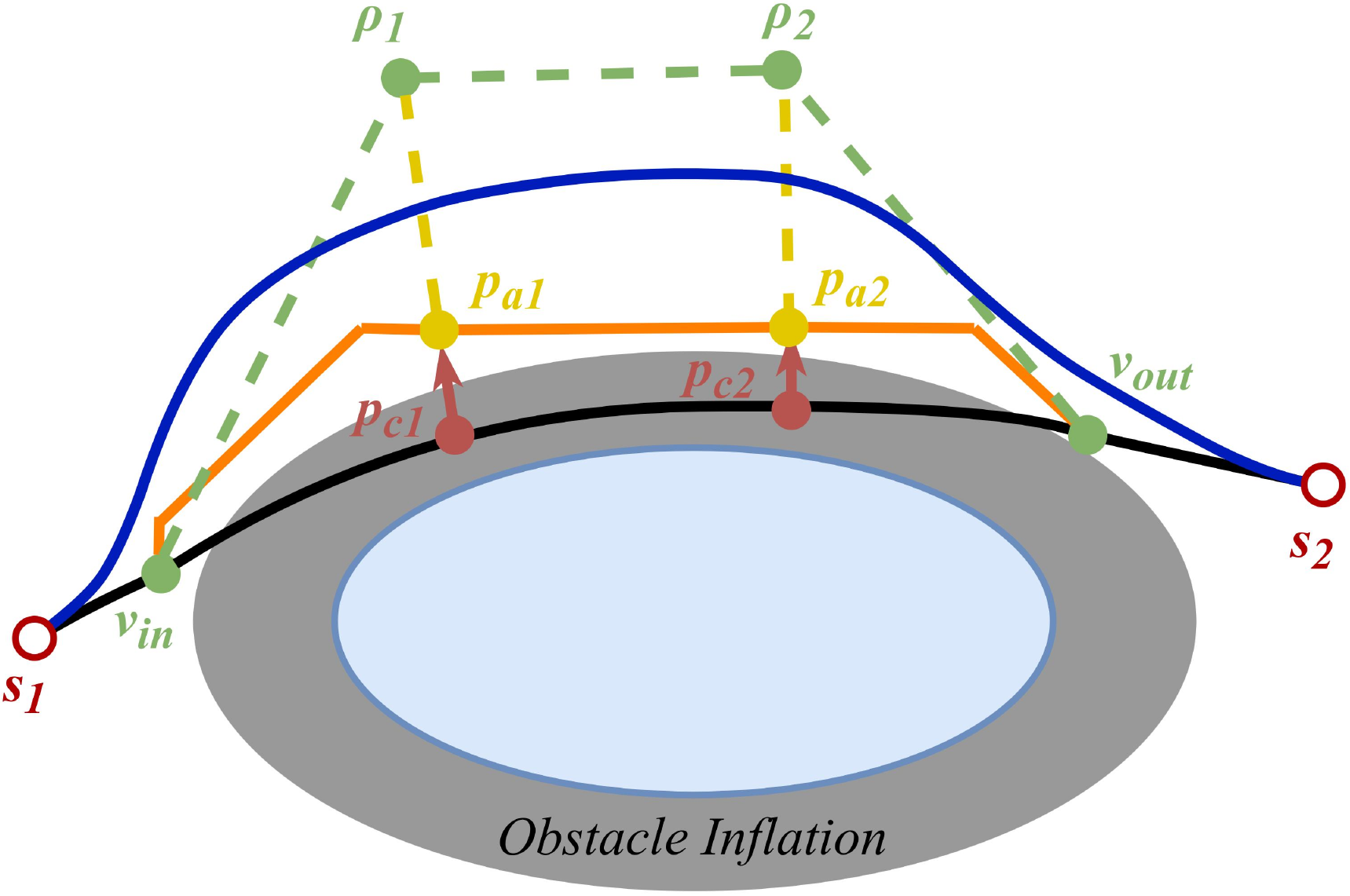}
	\caption{An illustration of trajectory optimization. The orange path is sampled by A$^*$ algorithm from $v_{in}$ to $v_{out}$, and then the anchor points $p_a$ are sampled. Finally, the blue path is generated by moving $p_a$ to a certain distance.}
	\label{fig:path_opti_illu}
\end{figure}

In order to optimize the trajectory, the constraints include the four following items, \textit{i.e.}, trajectory smoothness, collision-free, dynamic and original trajectory constraints. 
We divide the original trajectory into multiple segments based on time, so that there are multiple free states in the middle to prevent poor performance of whole trajectory optimization.
Then, the objective function is defined as
\begin{equation}~\label{eq:traj_opt}
	\mathcal{F} = \lambda_sf_s+\lambda_cf_c+\lambda_df_d+\lambda_of_o,
\end{equation}
where $f_s$, $f_c$, $f_d$ and $f_o$ denote the above items, and $\lambda_s$, $\lambda_c$, $\lambda_d$ and $\lambda_o$ are the weights respectively.

The trajectory smoothness term $f_s$ is the time integral of the trajectory jerk, \textit{i.e.},
\begin{equation}
	f_s = \int_{0}^{T}\|\bm{u}(t)\|^{2}dt=\bm{p}^{T} \bm{Q}_s \bm{p},
\end{equation}
where $\bm{p}^T=[\bm{p}_1^T, \bm{p}_2^T, ..., \bm{p}_{n_{seg}}^T]$ is composed of $n_{seg}$ trajectory segments, and $\bm{Q}_s$ is a diagonal matrix composed of time $T_i$ with different orders.

Referring to~\cite{zhou2020ego}, the collision-free term $f_c$ is described as the integral of the distance between the safety guidance point $\rho$ and the collision trajectory segment.
The process is illustrated in Fig.~\ref{fig:path_opti_illu}.
This makes the trajectory of the collision part tend to the safety guidance point, so as to keep away from obstacles.
\begin{equation}
	\begin{aligned}
			f_c &= \sum_{\rho\in\mathcal{P}_t}\int_{T_s^{\rho}}^{T_e^{\rho}}\|p(t)-p_{\rho}(t)\|^{2}dt\\
			&=\sum_{\rho\in\mathcal{P}_t}(\bm{p}-\bm{p}^{\rho})^{\top}\bm{Q}_{c,\rho}(\bm{p}-\bm{p}^{\rho}).
		\end{aligned}
\end{equation}
The collision part can be obtained by $\bf{FeasibilityCheck}()$, and updated using A$^*$ algorithm. $p_\rho(t)$ is the spatial location of $\rho$, and $T_s^{\rho}$, $T_e^{\rho}$ are the collision part influenced by $\rho$.

The dynamic constraint term $f_d$ limits the range of higher-order derivatives of the trajectory, and punishes the trajectory part that exceeds physical dynamics of USV. That is
\begin{equation}
	\begin{aligned}
			f_d &= \sum_{\iota\in\mathcal{S}}\int_{T_s^{\iota}}^{T_e^{\iota}}\|v(t)-v_{max}\|^{2}+\|a(t)-a_{max}\|^{2}dt\\
			&=\sum_{\iota\in\mathcal{S}}(\bm{p}-\bm{p}^{\iota})^{\top}\bm{Q}_{d,\iota}(\bm{p}-\bm{p}^{\iota}),
	\end{aligned}
\end{equation}
where $v_{max}$ and $a_{max}$ denote the maximal speed and acceleration of USV respectively.
$S$ is the trajectory part that exceeds the power limit.
$T_s^{\iota}$ and $T_e^{\iota}$ are the start and end time.

The original trajectory constraint term $f_o$ is similar to the collision-free term, except that the original trajectory position information $p^*(t)$ is used as the limit.
\begin{equation}
	f_o = \int_{0}^{T}\|p(t)-p^{*}(t)\|^{2}dt=(\bm{p}-\bm{p}^{*})^{T}\bm{Q}_o(\bm{p}-\bm{p}^{*}).
\end{equation}
The function of term is to constrain the safe part of original trajectory to maintain its original feasibility.

Referring to~\cite{richter2016polynomial}, trajectory optimization function~\ref{eq:traj_opt} has positive definite property, and the optimal solution can be obtained through a closed form.
If the trajectory still does not satisfy the requirements, multiple iterations are used to extract the unsatisfactory trajectory.
The weight of corresponding optimization term is modified with a greater impact, so that the trajectory is updated under new constraints.

In our framework, the trajectory generated by the front end based on dynamics is composed of several segments, so its overall smoothness can not be guaranteed, which is not optimal for tracking.
Therefore, these trajectory segments need to be optimized for trajectory smoothness.
We use the same optimization function as above, except that the smoothness weight $\lambda_s$ is enlarged and dynamic constraint weight $\lambda_d$ is reduced.
Meanwhile, because there is no collision in the optimized trajectory, we directly set the original trajectory as the safety route $\mathcal{P}_t$ to extract guidance point $\rho$ when constructing the collision-free term $f_c$.
The trajectory optimized by smoothing improves the feasibility and stability for USV.

\begin{figure*}[t]
	\centering
	\includegraphics[scale=0.55]{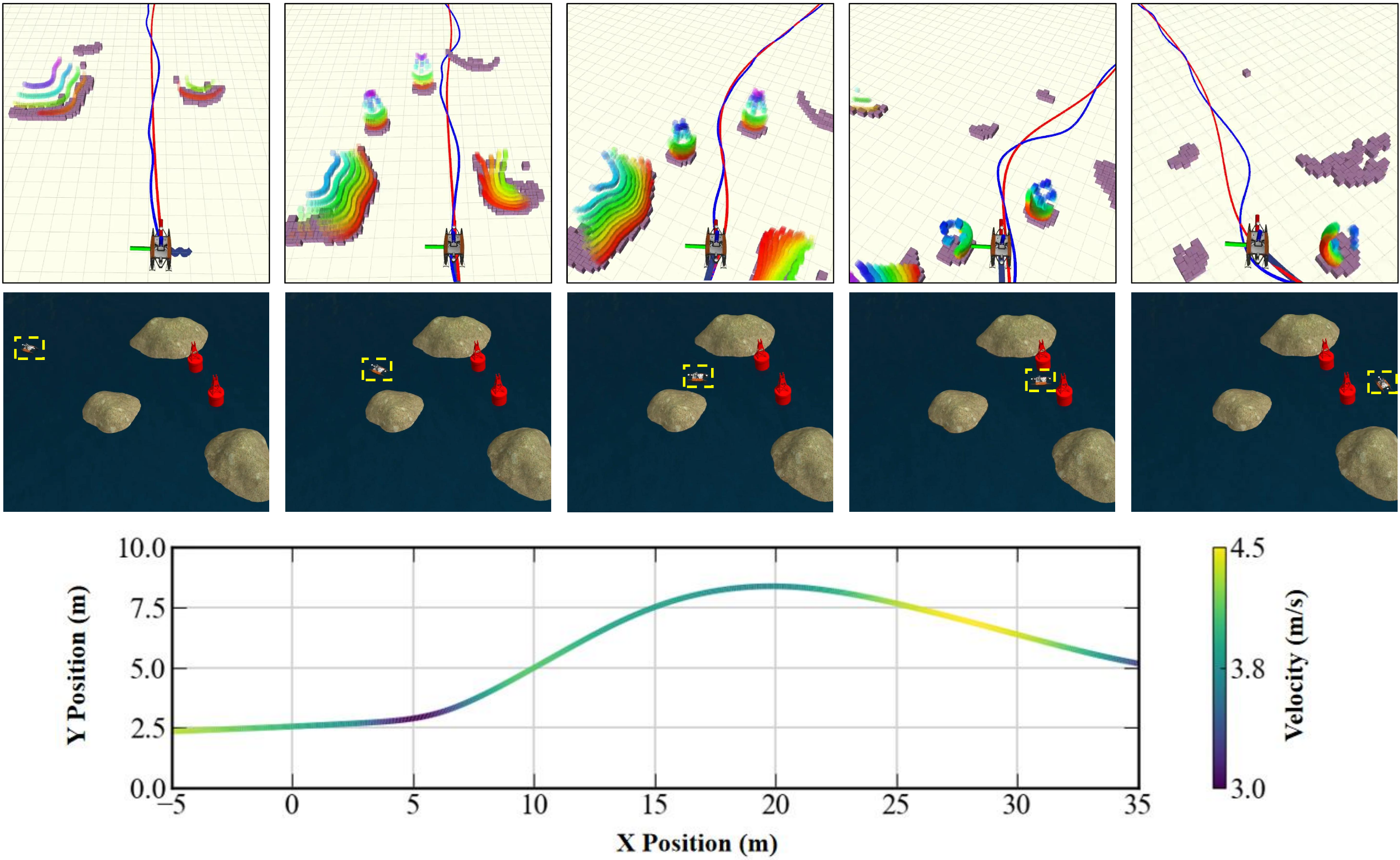}
	\caption{Visualization of trajectory planning over a short period of time with velocity profile.}
	\label{fig:path_planning}
\end{figure*}

\subsection{Motion tracking with NMPC}

The safe and feasible trajectory generated by the planning module needs effective trajectory tracking control to realize the autonomous navigation of USV.
To make USV have stable tracking ability in complex water environment, we set a high control frequency, \textit{i.e.}, 100Hz, and contain all the states $\bm{x}$ into the control law.

The nonlinear model predictive control (NMPC) method is selected as the controller.
According to the general form of MPC, a quadratic optimization problem is constructed about the error between the given trajectory state and the forward prediction state.
The following nonlinear discrete optimization problem is iteratively solved.
\begin{equation}
	\begin{gathered}
	\min _{\bm{u}} \bm{x}_{N}^{T} \boldsymbol{Q} \bm{x}_{N}+\sum_{i=0}^{N-1} \bm{x}_{i}^{T} \boldsymbol{Q} \bm{x}_{i}+\bm{u}_{i}^{T} \boldsymbol{R} \bm{u}_{i} \\
		s.t. \quad \bm{x}_{i+1}=\boldsymbol{f}_{RK4}\left(\bm{x}_{i}, \bm{u}_{i}, \Delta t\right) \\
		\bm{x}_{0}=\bm{x}_{init}, \bm{u}_{min} \leq \bm{u}_{i} \leq \bm{u}_{max}
	\end{gathered}
\end{equation}
where $\bm{x}_0$ is the actual state of USV at each solution cycle time, and the forward prediction time $T_f$ is discretized into $N$ time steps.
This optimization problem can be transformed into a sequential quadratic programming (SQP) problem to the extent of real-time iteration.
The control process is implemented through Casadi and Acados tool library~\cite{verschueren2022acados}.

\section{Experiment}

In this section, we perform simulation experiments using the open source USV simulator Otter~\cite{lenes2019autonomous} within the ROS environment.
The Otter USV is a catamaran cloned from the real world designed by the Maritime Robotics corporation.
It has a size of 2.0m long, 1.08m wide and 0.82m high with a total mass about 55kg.
The maximum velocity is about 2 knots.
Sensing equipments including lidar, camera and GPS have been installed in the simulator. 
The Otter USV sails in a rugged island environment with various floating obstacles.

The experimental scenario is constructed within Gazebo, and the scene size is about 300$\times$300 square meters.
The environment includes random known obstacles as well unknown obstacles.
Moreover, the scene contains a constant disturbance wind field and a surge that causes the USV attitude to swing.
The expansion of the barrier is set to 0.8m to prevent expansion.
All experiments are conducted on a laptop with an Intel i7-8700 CPU.

\subsection{Path planning analysis}

As shown in Fig.~\ref{fig:path_planning}, the Otter USV is travelling across a complex island environment with various unknown obstacles. The path planning module is able to generate a obstacle-avoidance trajectory in real-time.
The blue line shows the planning results by dynamic RRT$^*$, which is random and not optimal. After the optimization process proposed by our method, the USV is able to obtain a smooth and kinetic affine trajectory so as to cross all kinds of terrain in the best posture.

\begin{table}[t]
	\caption{Ablation study for the proposed method.}
	\centering
	\begin{tabular}{p{1.5cm}|p{0.6cm}p{0.6cm}p{0.6cm}p{0.6cm}p{0.6cm}p{0.6cm}}
	\hline
		\multirow{3}{*}{Method}  & Traj. Len.  & Traj. Time & Traj. Cost  & Node Util. & Algo. Time & Succ. Rate  \\
		     & ($m$)  &  ($s$)  & -   & ($\%$)  & ($ms$)   & ($\%$)    \\
	\hline
	\hline
		Baseline   & 71.4 & \textbf{20.5} & 24.3 & 29.7 & 58.1 & 86 \\
		+Local Opt. & 71.8 & 20.8 & 24.2 & 40.6 & 47.8 & 96   \\
		+Global Opt. & 69.6 & 21.6 & 22.1 & 30.8 & 60.2 & 84  \\
	\hline
		Proposed   & \textbf{69.3} & 21.3 & \textbf{21.9} & \textbf{41.8} & \textbf{45.6} & \textbf{98} \\
	\hline
	\end{tabular}
	\label{tab:ablation_study}
\end{table}

The ablation study for the proposed method is shown in Table~\ref{tab:ablation_study}, where the baseline denotes the dynamic RRT$^*$ algorithm, and local optimization (+Local Opt.) means trajectory local adjustment with differential flatness, and global optimization (+Global Opt.) means the trajectory optimization with dynamic constraints.
Trajectory length, time and cost are indicators that are associated with generated trajectory. Node utilization is the ratio of planning algorithm utilization to planned nodes and total sampling nodes, which is used to evaluate the efficiency of algorithm.
Algorithm time and successful rate are used to measure the performance of the algorithm.
The algorithm is tested repeatedly for 50 times, and a successful planning should be completed within 100$ms$.

In the baseline model test results, the node sampling rate is obviously low (29.7$\%$), and the problem of calculation timeout frequently occurs, resulting in a low success rate (86\%). At the same time, the algorithm consumes more time (58.1$ms$).
On the other hand, the length of trajectory generated by baseline model is long and the trajectory is not smooth. This also can be viewed from Fig.~\ref{fig:path_planning}.

After adding the local optimization method, the node utilization, planning time and successful rate have been significantly improved (40.6$\%$, 47.8$ms$, 96$\%$), which indicates that it has a significant effect on the efficiency and stability of the planning algorithm.
After the global optimization method is added, the path length and cost are reduced, because it makes the path smoother, and the algorithm time consumption does not increase too much.

As for the proposed method, we combine the baseline model with local and global optimization, resulting in a smoother trajectory and higher planning efficiency. The trajectory length and cost are the lowest (69.3$m$, 21.9) while the node utilization, algorithm time and successful rate achieve the best (41.8$\%$, 45.6$ms$, 98$\%$).

\begin{figure}[t]
	\centering
	\includegraphics[scale=0.45]{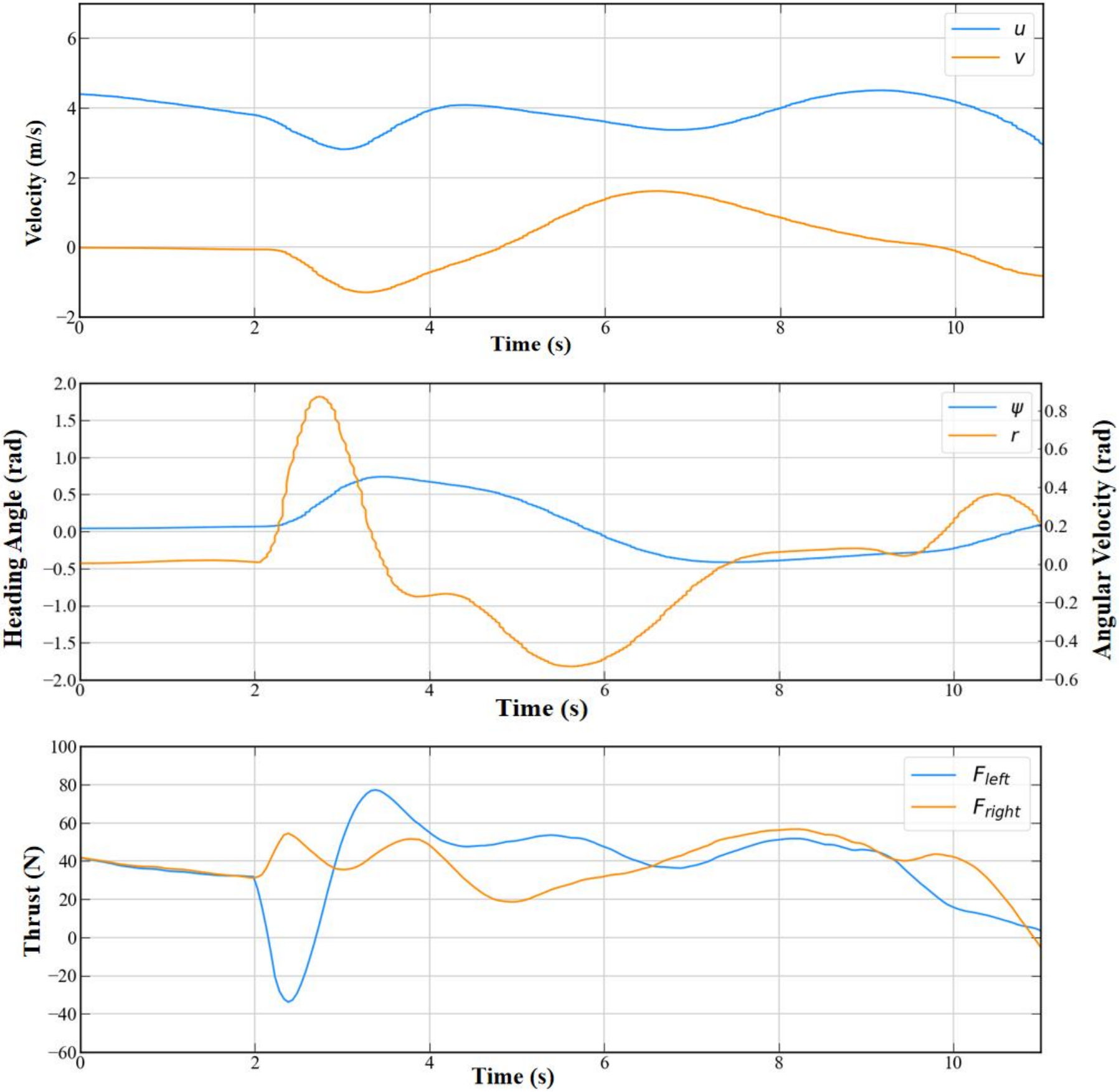}
	\caption{State variables with time for the planned trajectory.}
	\label{fig:state_variable}
\end{figure}

\begin{figure*}[t]
	\centering
	\includegraphics[scale=0.59]{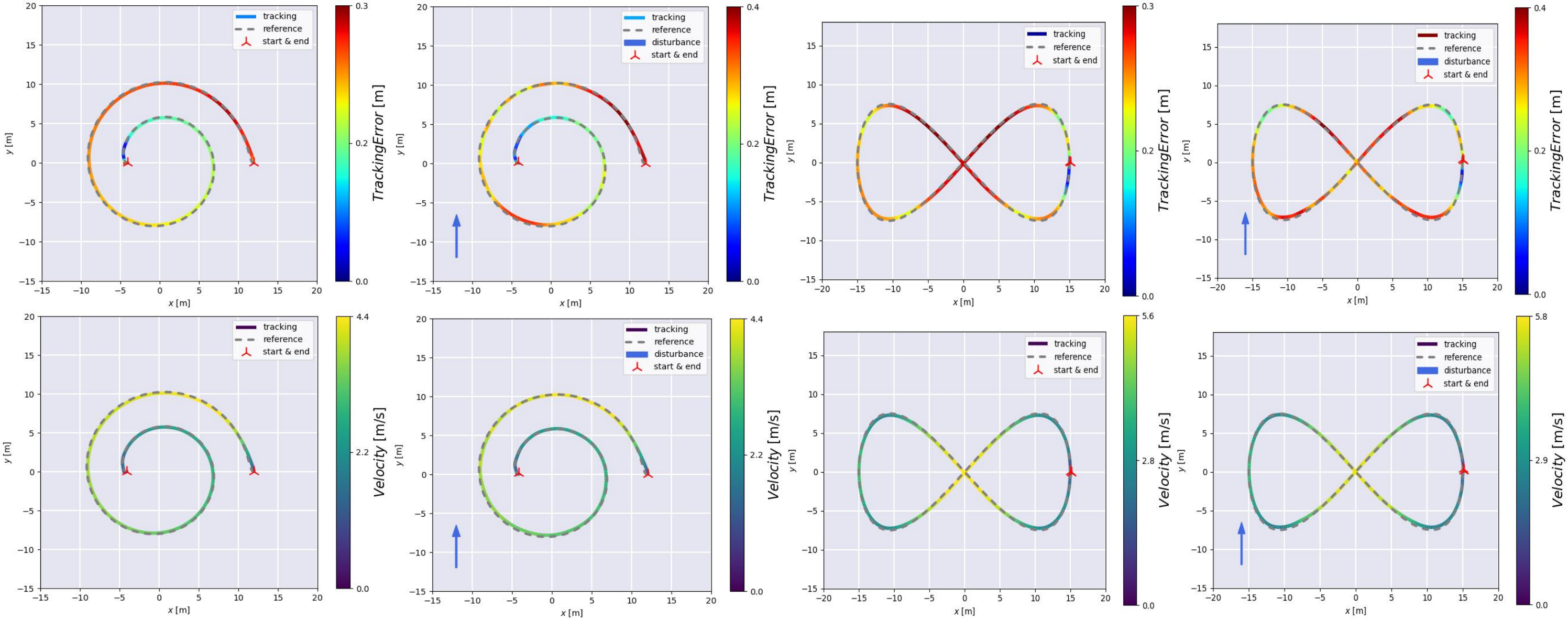}
	\caption{Visualization of motion tracking results of a spiral type and splayed shape trajectory with and without disturbances.}
	\label{fig:vis_tracking}
\end{figure*}

Taking a closer look at the state variables of the planned trajectory as shown in Fig.~\ref{fig:state_variable}, there are few shocks in the surge and sway velocity curves, as well as the yaw and its rate.
As for the control inputs, the desired thrusts are generally stable, which shows that the control performance requirements for the USV are relatively low in this case.

\subsection{Motion control analysis}

As for the motion control performance, the results are illustrated in Fig.~\ref{fig:vis_tracking}.
Two types of trajectory are planned to be tracked with the proposed control method, including the spiral type and splayed shape trajectory.
Meanwhile, we test the control performance with and without disturbance.
From the results, we can see that the tracking control performance remains a high level.
Overall, the real trajectory of USV is consistent with the planned trajectory, and the tracking error is controlled within 0.4m, in a 300$\times$300 square meters map.
The tracking velocity is quite large, ranging up to about 4m/s, which is quite a large velocity with regard to an USV.

\begin{table}[t]
	\caption{Ablation study for differential flatness (DF) in motion control.}
	\centering
	\begin{tabular}{p{1.4cm}|p{0.7cm}p{0.7cm}p{0.7cm}p{0.7cm}p{1.1cm}}
	\hline
		\multirow{3}{*}{Method}  & Mean Error  & Max Error & Mean Vel.  & Max Vel. & Ang. Vel. Integral   \\
		     & ($m$)  &  ($m$)  & ($m/s$)   & ($m/s$)  & ($rad^2/s$)    \\
	\hline
	\hline
		w.o DF   & 0.257 & 0.768 & 2.93 & 4.31 & 2.01  \\
		Ours   & \textbf{0.121} & \textbf{0.483} & \textbf{3.32} & \textbf{4.37} & \textbf{0.95}  \\
	\hline
	\end{tabular}
	\label{tab:control_comparison}
\end{table}

The quantitative comparisons between the baseline model (NMPC without DF) and the proposed method are summarized in Table~\ref{tab:control_comparison}.
Without differential flatness, although the USV trajectory tracking algorithm can still track the trajectory in real time, the control effect is poor and the navigation is not smooth, and the mean tracking error achieves 0.257$m$.
After adding differential flatness, the angular velocity integral has significantly decreased (from 1.97 to 0.82), which indicates that differential flatness makes a better connection between path planning and motion control components.
It has a better improvement on the robustness of motion control.

\section{Conclusion}
In this paper, we propose a novel yet efficient trajectory planning and motion control algorithm for USV.
Traditional methods use trajectory designed for UAV or AGV as the guidance to generate obstacle-avoidance path, which ignores the inherent under-actuated characteristics of USV.
Based on this, we propose a trajectory planning algorithm that concerns the hull dynamics during the process, and use differential flatness to improve algorithm efficiency. The generated trajectory is in good agreement with the characteristics of the hull, so the control effect remains a high level.
We perform extensive simulation experiments to verify the effectiveness of the proposed method in planning and control performance for USV.
In the future, we will carry out high-precision system identification on the real ship and realize it in real objects.

\bibliographystyle{IEEEtran}
\bibliography{refs}

\end{document}